\documentclass{article}

    \PassOptionsToPackage{numbers, compress}{natbib}

 \usepackage[main, final]{neurips_2026}

\usepackage[utf8]{inputenc} 
\usepackage[T1]{fontenc}    
\usepackage{hyperref}       
\usepackage{url}            
\usepackage{booktabs}       
\usepackage{amsfonts}       
\usepackage{nicefrac}       
\usepackage{microtype}      
\usepackage{xcolor}         
\usepackage{graphicx}
\usepackage{amsmath}
\usepackage{multirow}
\usepackage{makecell}
\usepackage{caption}
\usepackage{colortbl}
\usepackage{tcolorbox}
\usepackage[tikz]{bclogo}
\usepackage{subfigure}
\usepackage{subcaption}
\usepackage{arydshln}
\usepackage{float}
\usepackage{pifont}
\usepackage{alltt}
\usepackage{dblfloatfix}
\usepackage{enumitem}
\usepackage{wrapfig}
\usepackage{adjustbox}
\usepackage{tabularx}
\usepackage{amssymb}
\title{OmniSimulator: Aligning Small Language Models \\ for Authentic Heterogeneous Behavior Modeling}

\author{
Jiawei Chen$^{1,2}$,
Ruoxi Xu$^{1,2}$,
Boxi Cao$^1$,
Zhuoqun Li$^{1,2}$,
Ruotong Pan$^3$,
Yunfei Zhang$^3$,\\
\textbf{Zhirui Yang}$^3$,
\textbf{Jingyan Chen}$^3$,
\textbf{Tingting Gao}$^3$,
\textbf{Han Li}$^3$,
\textbf{Yaojie Lu}$^1$,\\
\textbf{Xianpei Han}$^1$,
\textbf{Le Sun}$^1$,
\textbf{Xiangyu Wu}$^3$\setcounter{footnote}{1}\thanks{Corresponding authors.}\hspace{0.5em},
\textbf{Hongyu Lin}$^1$\footnotemark[2]\hspace{0.5em}\\
$^1$ Chinese Information Processing Laboratory, Institute of Software, \\
  Chinese Academy of Sciences \\
$^2$ University of Chinese Academy of Sciences \\
$^3$ Kuaishou Technology \\
\texttt{chenjiawei2024@iscas.ac.cn}
} 
\hypersetup{
    colorlinks=true,
    linkcolor=black,
    citecolor=black,
    urlcolor=blue
}

\tcbuselibrary{skins}
\definecolor{msftBlack}{RGB}{0,0,0}
\newtcolorbox{findingBox}{
    enhanced,
    colback=msftBlack!05,
    colframe=msftBlack!10,
    arc=1mm,
    boxrule=0.5pt,
    left=2mm, right=2mm, top=2mm, bottom=2mm,
    drop fuzzy shadow,
    fontupper=\em,
    notitle
}

\begin{document}

\maketitle

\setcounter{footnote}{0}

\begin{abstract}
Large language models (LLMs) have shown great potential as general-purpose user simulators for interactive systems. Prior studies have demonstrated that training on high-quality data can substantially improve model reasoning and downstream task performance. However, the upper bound of user simulation is still constrained by the model's reasoning paradigm: native chain-of-thought often degenerates into shallow pattern matching and fails to adequately model the behavioral factors underlying real user decisions. In this paper, we propose \textbf{\textit{OmniSimulator}}, a novel training framework designed to improve the predictive ability of small LLMs for realistic user simulation on short-video platform environments. Specifically, OmniSimulator organizes behavior-oriented reasoning through a structured schema with five dimensions, carefully selects data from real user-behavior scenarios, and introduces a teacher model to generate high-quality reasoning traces under the proposed schema, helping small models learn behavior-relevant reasoning patterns rather than relying on superficial behavioral correlations. Based on this process, we construct 9,881 high-quality training instances from three months of historical interaction trajectories of 200 real users, averaging about 50 distilled examples per user. Experimental results show that OmniSimulator improves the average score of Qwen3-8B by 34.33\%, achieves performance competitive with Claude-Sonnet-4.5, and supports smaller-scale, self-hosted deployment. 
These results suggest that learning structured behavior-oriented reasoning patterns is a promising direction towards scalable and high-fidelity LLM-based user simulation\footnote{Our code is publicly available at \url{https://github.com/icip-cas/OmniSimulator}}.
\end{abstract}

\section{Introduction}

Large language models (LLMs) have recently shown strong promise in a growing range of downstream applications \cite{zhang2020evaluating,jin2025search,sun2025llm,huang2024understanding,chen2025consistentchat}, including recommender systems \cite{bernardi2021simulations,geng2022recommendation,bougie2025simuser}, advertising \cite{reisenbichler2026applying}, customer service \cite{wulf2024exploring,li2025performance}, and interactive agents \cite{yao2022react,wu2024autogen}. Beyond their generative ability, an increasingly appealing direction is to use LLMs as \emph{general-purpose user simulators} \cite{argyle2023out,park2023generative,bougie2025simuser}, enabling offline evaluation, policy testing, product optimization, and behavior modeling at a scale that would otherwise require expensive online experiments \cite{park2022social,park2023generative}. This trend is especially important for short-video platforms, where user behavior is inherently long-horizon, cross-scenario, and heterogeneous \cite{chen2026towards}: a user may browse videos, enter live rooms, click ads, shop in e-commerce, search for information, and interact with customer service within one unified behavioral trajectory. Recent studies have shown that training on high-quality data can substantially improve reasoning ability and downstream task performance of LLMs, suggesting that post-training and supervision quality are critical for building stronger simulators. However, real user simulation remains far from solved. On challenging real-world benchmarks such as OmniBehavior~\cite{chen2026towards}, even frontier models achieve limited performance, while also exhibiting systematic biases such as overestimating positive actions, generating overly optimistic utterances, and collapsing different users into an average persona \cite{bian2025social,chen2026towards}. Moreover, although closed-source models such as Claude-Opus-4.5 \cite{anthropic2025claudeopus45} provide stronger performance, their cost becomes prohibitive when conditioning on long interaction histories, making them difficult to deploy at industrial scale.

A difficulty of user simulation is that real behavior is not generated by direct surface matching between historical interactions and the current scenario. Instead, user actions are mediated by latent mental states, such as stable preference, recent interest drift, transient fatigue, emotional tendency, and momentary decision style. In other words, predicting whether a user will click, purchase, continue watching, or say a particular utterance requires the model to first recover \emph{why this user is likely to act in that way at the current scenario}. This observation naturally suggests that reasoning capability counts. If the model can explicitly infer the hidden state underlying the next action, it should in principle make more faithful predictions than a model that only performs shallow behavioral imitation. Chain-of-thought (CoT) therefore appears to be a natural candidate for improving user simulation, since it provides an intermediate reasoning process between observed history and predicted behavior. However, our empirical analysis reveals that simply enabling native thinking mode brings almost no benefit for user simulation: across Qwen3 models, the performance of thinking and non-thinking settings is often nearly identical, and in some cases the thinking variant even performs worse. A closer inspection of native CoT outputs explains why: existing models rarely use reasoning to reconstruct the user's latent mental state; instead, they tend to produce generic justifications, shallow pattern matching, or direct guesses based on superficial correlations. This suggests that the bottleneck is not merely whether the model generates more reasoning tokens, but whether its reasoning follows a behavior-oriented structure.

\begin{figure*}[t]
\centering
\includegraphics[width=\textwidth]{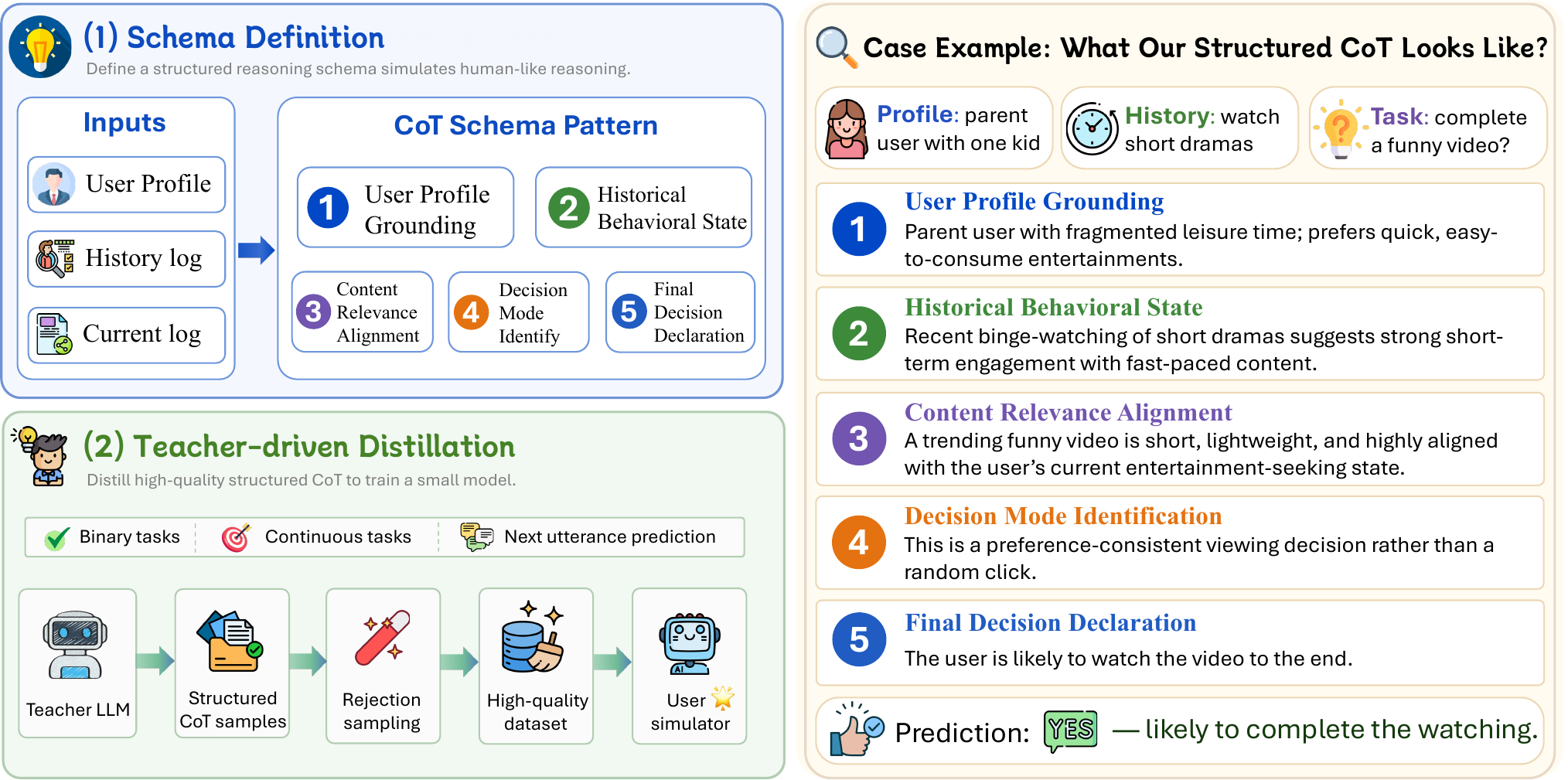}
\caption{\textbf{Overview of OmniSimulator.} The framework consists of two stages. \textbf{(1) Schema Definition:} we define a structured chain-of-thought schema that guides the model to simulate human-like internal reasoning from three inputs: user profile, historical interaction logs, and the current scenario. The schema decomposes reasoning into five dimensions: \emph{User Profile Grounding}, \emph{Historical Behavioral State}, \emph{Content Relevance Alignment}, \emph{Decision Mode Identification}, and \emph{Final Decision Declaration}. \textbf{(2) Teacher-driven Distillation:} a teacher LLM generates structured reasoning traces under the schema for different task types, and high-quality traces are selected through rejection sampling to construct the distilled training set for small user simulators. The right panel shows a case example of the structured user-state reasoning trace, where the model infers latent user state step by step and then predicts the final correct behavior.}
\label{fig:framework}
\end{figure*}

Motivated by this, we propose \textbf{OmniSimulator}, a novel training framework designed to improve small LLMs for real user simulation on short-video platform environments. The key idea is to teach the model to follow behavior-oriented structured reasoning rather than free-form reasoning. Specifically, OmniSimulator introduces a structured CoT schema that forces the model to infer latent user states through five dimensions: \emph{User Profile Grounding}, \emph{Historical Behavioral State Inference}, \emph{Content Relevance Alignment}, \emph{Decision Mode Identification}, and \emph{Final Decision Declaration}. On top of this schema, we build a task-aware causal distillation pipeline: a teacher model (GPT-OSS-120B~\citep{agarwal2025gpt}) generates multiple reasoning traces for each prediction point under task-specific prompting strategies, and only traces consistent with ground truth are retained through rejection sampling. Following this process, we construct 9,881 high-quality distilled training instances from three months of interaction histories of 200 real users. 
Experiments show that OmniSimulator substantially improves small models over zero-shot and free-CoT baselines, enables Qwen3-8B to achieve performance competitive with strong proprietary models such as Claude-Sonnet-4.5 \cite{anthropic2025claudesonnet45}, and alleviates overestimation bias while supporting smaller-scale deployment. These results suggest that the upper bound of LLM-based user simulation is constrained not only by model scale, but also by the reasoning supervision paradigm. 

Our main contributions are summarized as follows:
\begin{itemize}[leftmargin=2em, itemsep=1pt, topsep=0pt]
    \item We identify a key weakness of existing LLM-based user simulators: native CoT struggles to predict realistic user behavior, yielding only marginal improvements in behavioral prediction over non-thinking settings.
    \item We propose OmniSimulator, a novel training framework that combines a structured five-dimensional CoT schema with a task-aware causal distillation pipeline, enabling small models to learn behavior-oriented reasoning patterns for user simulation.
    \item We construct 9,881 high-quality distilled training instances from real-world long-horizon user trajectories, and OmniSimulator substantially improves small LLMs, achieves performance competitive with strong closed-source baselines and mitigates overestimation bias while supporting efficient small-model deployment.
\end{itemize}

\section{Methodology}
\label{sec:method}

In this section, we first formulate user behavior simulation on short-video platforms, and then present \textbf{OmniSimulator}, our training framework for improving small LLMs on realistic user simulation. The central idea is that real user behavior is not generated by direct surface matching between historical interactions and the current scenario. Instead, it is mediated by a latent decision process shaped by stable preference, recent interest drift, transient affect, and the decision regime at the current moment. Existing LLMs, even with native thinking mode, often fail to reconstruct this hidden process and instead rely on shallow behavioral priors. To address this issue, OmniSimulator explicitly teaches the model to infer latent user states before predicting behavior. The framework consists of two tightly coupled components: (1) a \emph{structured chain-of-thought schema} that organizes reasoning into a fixed sequence of human-like internal deliberation steps, and (2) a \emph{task-aware rationale distillation pipeline} that converts teacher-generated reasoning traces into high-quality supervision for student models.

\subsection{Problem Formulation}
\label{subsec:problem_formulation}

We formulate user simulation as a user-conditioned prediction problem over long-horizon, cross-scenario interaction trajectories. For a user $u$, let
\begin{equation}
H_t^u = \{e_1^u, e_2^u, \dots, e_t^u\}
\label{eq:history}
\end{equation}
denote the historical interaction sequence before time step $t+1$, where each event $e_i^u$ may come from different scenarios such as video browsing, live streaming, e-commerce shopping, customer-service dialogue, advertisement recommendation, and search. Let $x_{t+1}^u$ denote the current scenario context. The goal is to predict the corresponding user response
\begin{equation}
y_{t+1}^u = f_\theta(H_t^u, x_{t+1}^u),
\label{eq:prediction}
\end{equation}
where $y_{t+1}^u$ may be a binary decision, a continuous value, or a natural-language utterance.

A direct mapping from $(H_t^u, x_{t+1}^u)$ to $y_{t+1}^u$ is often insufficient. In realistic platforms, a user’s next behavior is mediated by hidden mental states that are not explicitly recorded in logs, such as stable preference, short-term tendency, fatigue, emotion, and momentary decision style. We therefore introduce an implicit latent state $z_t^u$ and decompose user simulation as
\begin{equation}
p(y_{t+1}^u \mid H_t^u, x_{t+1}^u)
=
\sum_{z_t^u}
p(y_{t+1}^u \mid z_t^u, x_{t+1}^u)\,
p(z_t^u \mid H_t^u).
\label{eq:latent}
\end{equation}
Since $z_t^u$ is unobserved, OmniSimulator does not supervise it as an explicit label. Instead, we represent it as a structured reasoning trace in natural language, and train the model to infer this trace before answering the final behavior prediction.

\subsection{Preliminary Analysis: Revealing Native Thinking Effects}
\label{subsec:preliminary_analysis}

\begin{wrapfigure}[15]{r}{0.5\textwidth}
\vspace{-1.0em}
\centering
\includegraphics[width=\linewidth]{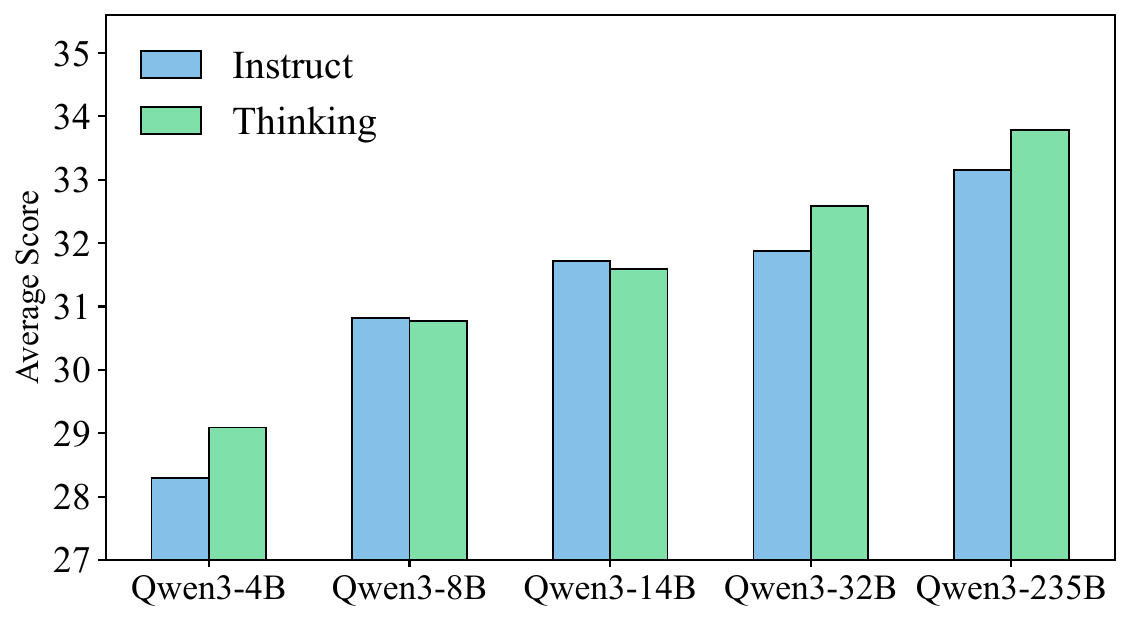}
\caption{Preliminary analysis on native thinking mode. Across Qwen backbones, enabling native thinking yields only marginal or inconsistent gains over standard instruction-following inference.}
\label{fig:preliminary_qwen_analysis}
\end{wrapfigure}

We first investigate whether native thinking mode alone is sufficient for realistic user simulation. Intuitively, this task appears to benefit from reasoning, since the model must recover the hidden factors that connect long interaction histories to current decisions. However, our preliminary analysis reveals that this intuition does not hold under standard prompting.

Figure~\ref{fig:preliminary_qwen_analysis} compares the average OmniBehavior scores of multiple Qwen3-series \cite{yang2025qwen3} backbones under two settings: \textsc{Instruct} and \textsc{Thinking}. We observe that enabling native thinking yields little or no consistent gain over standard instruction-following inference. Across model scales, the two settings are often nearly identical, and in some cases the thinking variant even performs worse. This result suggests that the bottleneck is not the absence of additional reasoning tokens, but the absence of the \emph{right reasoning structure}. By inspecting native CoT outputs, we find that models often produce generic justifications or shallow guesses, failing to explicitly reconstruct the user’s latent psychological state. This observation motivates our approach: rather than simply allowing the model to think more, we guide it to follow a structured, behavior-oriented reasoning process.

\subsection{Overview of OmniSimulator}
\label{subsec:overview}

Figure~\ref{fig:framework} illustrates the overall framework. For each prediction point, we first provide the teacher model with the user’s historical interaction trajectory and the current scenario context. The teacher is required to reason under a structured schema that explicitly reconstructs the user’s latent mental state. To increase diversity, we sample multiple reasoning traces for each example with stochastic decoding. Depending on the task type, we apply different prompting strategies to control the degree of access to ground truth. The resulting traces are filtered by rejection sampling, and only those whose final answers are consistent with the observed outcome are retained as supervised fine-tuning data. Finally, the student model is trained to generate both the structured reasoning trace and the final prediction.

This design follows a simple principle: rather than only teaching the model what to predict, we also provide structured rationale supervision for how the prediction may be derived from behavioral evidence. In this way, latent-state reconstruction becomes explicit, learnable, and distillable.

\paragraph{Structured Chain-of-Thought Schema.}
\label{para:structured_cot}
The first stage of OmniSimulator is a structured CoT schema tailored for user simulation. Our design is motivated not only by the empirical analysis above, but also by cognitive theories of human behavior. Prior work suggests that behavior is mediated by stable person-level characteristics, dynamically activated cognitive-affective states, and their interaction with situational cues~\cite{mischel1995cognitive}. In parallel, dual-process accounts further distinguish between reflective, deliberative decision making and more impulsive behavior pathways~\cite{strack2004reflective}. Native thinking mode alone is therefore insufficient, because it leaves this behavioral reasoning path under-specified for human behavior simulation tasks. To address this issue, we require the model to generate reasoning within a \texttt{<think>} block following five fixed dimensions:

\begin{enumerate}[leftmargin=2em, itemsep=1pt, topsep=0pt]
    \item \textbf{User Profile Grounding:} The model first interprets stable user characteristics reflected in long-term history, such as age, gender, educational background. This step provides the global behavioral background of the user.
    \item \textbf{Historical Behavioral State:} The model then infers the user’s dynamic internal state from recent and long-horizon interactions. This includes long-term interest continuation, short-term preference drift, current fatigue level, and emotional tendency. Compared with the previous step, the focus here is not the user's general profile, but the user’s state from the history sequences at the current moment.
    \item \textbf{Content Relevance Alignment:} Next, the model matches the inferred user state with the current scenario and content. This step examines whether the current item is aligned with the user’s active interests, intent, mood, or aversion, and serves as the bridge between historical state and prediction item.
    \item \textbf{Decision Mode Identification:} The model further determines whether the current behavior is primarily driven by \emph{rational decision} or \emph{random decision}. Rational decisions are mainly explained by interpretable factors such as fatigue levels in this session, while random decisions are more influenced by curiosity or stochasticity. This distinction is essential for modeling negative outcomes and low-confidence actions.
    \item \textbf{Final Decision Declaration:} Before closing the \texttt{</think>} block, the model should explicitly state its final decision, and this declaration is required to be fully consistent with the final output answer. This design reduces reasoning-answer mismatch and improves the controllability of the generated traces.
\end{enumerate}

With this schema, OmniSimulator decomposes user simulation into the following process:
\begin{equation}
(H_t^u, x_{t+1}^u) \rightarrow r_t^u \rightarrow y_{t+1}^u,
\label{eq:trace}
\end{equation}
where $r_t^u$ denotes the structured reasoning trace. In practice, $r_t^u$ serves as a textual proxy of the unobserved latent state $z_t^u$. Compared with unconstrained CoT, this schema forces the model to move from observable behaviors to hidden user states, and then from hidden user states to final actions.

\paragraph{Task-aware Rationale Distillation.}
\label{para:distillation}

Structured reasoning alone is not sufficient if teacher-generated traces are noisy. We therefore design a task-aware rationale distillation pipeline to obtain high-quality supervision for student training.

For each prediction item, we use a teacher model (GPT-OSS-120B \cite{agarwal2025gpt} in our experiment) to rollout $N$ candidate reasoning traces under the structured schema. Formally, given an example $(H_t^u, x_{t+1}^u, y_{t+1}^u)$, we sample
\begin{equation}
\mathcal{R}_t^u = \{r_{t,1}^u, r_{t,2}^u, \dots, r_{t,N}^u\}
\label{eq:samples}
\end{equation}
with relatively high temperature. This design allows the teacher to generate multiple candidate rationales for each prediction instance. However, different task types have different output spaces and different risks of information leakage, so we use different prompting strategies for different tasks.

\begin{itemize}[leftmargin=2em, itemsep=1pt, topsep=0pt]
    \item \textbf{Binary tasks:} For binary prediction, we use \emph{forward-only reasoning}. The teacher is not given the ground-truth label, and must predict the outcome solely from historical evidence and the current scenario. This avoids hindsight bias, since directly injecting the correct label would encourage the teacher to rationalize the answer after the fact.
    \item \textbf{Continuous-value tasks:} For continuous prediction, exact-value sampling is more difficult because the target space is broad and numerically sensitive. Pure forward reasoning often yields semantically reasonable but poorly calibrated values. To stabilize teacher generation, we provide a coarse range hint derived from the ground truth, rather than the exact answer itself. This provides task-specific guidance for generating candidate rationales with more appropriate numerical scales.
    \item \textbf{Text-generation tasks:} For textual simulation, namely next utterance prediction, the answer space is open-ended and highly diverse. Unconstrained multi-sample generation is therefore both costly and unstable. We directly inject the ground-truth utterance and require the teacher’s reasoning to converge to this answer. In this setting, the role of the reasoning trace is to explain why the observed utterance is consistent with the available behavioral history and the current conversational context.
\end{itemize}

After teacher generation, we apply rejection sampling and retain only those traces whose final prediction is consistent with ground truth:
\begin{equation}
\widetilde{\mathcal{R}}_t^u
=
\{r \in \mathcal{R}_t^u \mid \mathrm{Ans}(r)=y_{t+1}^u\},
\label{eq:rejection}
\end{equation}
where $\mathrm{Ans}(r)$ denotes the final answer implied by trace $r$. For each instance with accepted candidates, we select the trace with the highest teacher mean token log-probability, breaking ties by shorter length, to construct the supervised training data. If none of the sampled traces matches the observed outcome, the example is excluded from the training set and stored separately for analysis. In practice, this mainly occurs on difficult binary examples. This design deliberately prioritizes supervision quality over raw sample coverage.

\paragraph{Student Training.}
\label{para:training}

Based on the filtered reasoning traces, we construct the distilled supervised fine-tuning dataset
\begin{equation}
\mathcal{D}_{\mathrm{SFT}}
=
\{(H_t^u, x_{t+1}^u, r_t^u, y_{t+1}^u)\}.
\label{eq:dataset}
\end{equation}
The student model is trained to generate both the structured reasoning trace and the final answer:
\begin{equation}
p_\theta(r_t^u, y_{t+1}^u \mid H_t^u, x_{t+1}^u).
\label{eq:student}
\end{equation}
We optimize the standard autoregressive negative log-likelihood:
\begin{equation}
\mathcal{L}_{\mathrm{SFT}}
=
-
\sum_{(H,x,r,y)\in \mathcal{D}_{\mathrm{SFT}}}
\log p_\theta(r,y \mid H,x).
\label{eq:loss}
\end{equation}

This objective differs from direct answer-only supervision in an important way. Instead of only optimizing the final prediction, it encourages the student model to learn an internal decision process that first reconstructs the user’s latent state and then derives behavior from it. As a result, the student is trained to imitate human-like deliberation rather than only mimic surface-level action patterns.

The student model receives the user profile, historical interaction sequence, and current scenario context. It then generates a structured reasoning trace under the predefined schema and outputs the final prediction. No manually constructed external annotation, or additional structured memory is required. All latent-state reconstruction is performed directly from the observed trajectory.
This design preserves the end-to-end simplicity of LLM-based user simulation while improving faithfulness to real behavior. In particular, because the model is explicitly trained to reason about latent mental states, it is better positioned to capture negative feedback, temporary preference shifts, and user-specific heterogeneity, which are often suppressed in standard simulation pipelines.

\section{Experiments}
\label{sec:experiments}
We conduct experiments on realistic multi-scenario user simulation tasks to address three research questions: Q1: How effectively does OmniSimulator improve small LLMs across diverse user behaviors and scenarios? Q2: How competitive are OmniSimulator-trained small LLMs compared with frontier LLMs and human baselines? Q3: Which design choices in the reasoning schema and training pipeline contribute most to the performance gains?

\subsection{Experimental Setup}
\label{subsec:exp_setup}

\paragraph{Benchmark.}
We conduct experiments on OmniBehavior \cite{chen2026towards}, a real-world benchmark for LLM-based user simulation built from long-horizon, cross-scenario, and heterogeneous behavior traces. The benchmark is constructed from three months of complete interaction trajectories of 200 representative users on a short-video platform, covering video browsing, live streaming, advertisement, e-commerce (including customer service dialogues), and search scenarios. Prior analysis on OmniBehavior shows that realistic user decisions are inherently long-horizon and cross-scenario, and that current frontier LLMs still exhibit a substantial capability gap on this benchmark.

\paragraph{Training data construction.}
Our training data is distilled from the same realistic platform setting as OmniBehavior, but from users and trajectories excluded from the benchmark. Specifically, we sample another 200 representative users from Kuaishou platform and three-month collection window, with no overlap in users, trajectories, or target behaviors with the benchmark. The split is user-disjoint but temporally overlapping, with no global timestamp cutoff; each prediction instance contains only the corresponding user's interactions before its target time. From these multi-scenario historical trajectories, we construct prediction instances by pairing each target behavior with its preceding interaction history under a 32K context. We then apply the pipeline described in Section~\ref{sec:method}: a teacher model rollouts multiple reasoning traces under our structured schema, and only the traces whose final predictions are consistent with ground truth are retained. Following this process, we obtain 9,881 high-quality distilled training instances, which serve as supervision for student model fine-tuning.

\paragraph{Models and baselines.}
We evaluate four student backbones: Qwen3-0.6B, Qwen3-1.7B, Qwen3-4B, and Qwen3-8B. For each model, we compare four settings:
(1) Zero-shot (w/o think), directly predicts behaviors under standard instruction-following inference;
(2) Zero-shot (w/ think), enables the model's native thinking mode;
(3) SFT w/ free CoT, which fine-tunes the model on teacher-generated reasoning traces without our structured schema; and
(4) OmniSimulator, our full method with structured CoT and task-aware causal distillation.
Additional training variants are evaluated on Qwen3-8B to examine the effects of reasoning supervision, rationale design, and rejection sampling.

\paragraph{Additional Training Controls.}
To further examine the contributions of reasoning supervision and rejection sampling, we conduct additional comparisons on Qwen3-8B. These include Answer-only SFT using the same retained training examples, Synthetic Rationale SFT \citep{lu2026can}, and OmniSimulator without rejection sampling (RS). The results are presented in Section~\ref{sec:training_strategy}.

\paragraph{Metrics.}
Following OmniBehavior, we categorize user behaviors into 3 types.
For binary behaviors (e.g., like, share), we formulate simulation as a classification problem and report the F1-score.
For continuous behaviors (watch duration), standard absolute error introduces bias toward longer videos. 
To mitigate this, we report a normalized score defined as \(100 \times (1-\mathrm{NMAE})\), where NMAE normalizes the prediction error by the video's duration and higher scores indicate better performance.
Finally, for textual behaviors (e.g., customer service dialog), we utilize Claude-Sonnet-4.5 as judge to quantify the performance of the simulator which measures four criteria (intent fidelity, persona mimicry, knowledge boundary, and semantic alignment). We further validate the textual evaluation using two additional LLM judges and a blinded paired human evaluation, with detailed results provided in the appendix.

\begin{table*}[t]
\captionsetup{belowskip=0pt, aboveskip=6pt}
\caption{Comprehensive comparison of different training settings on the OmniBehavior Benchmark. We categorize user behaviors into three types: binary behaviors, continuous behaviors, and textual behaviors. The average score represents the aggregated performance.}
\resizebox{\textwidth}{!}{
\begin{tabular}{>{\centering\arraybackslash}m{1.6cm}lccccccc}
\toprule
\multirow{2}{*}{Models} & \multirow{2}{*}{Settings} & \multicolumn{2}{c}{Video} & Live & Ads & \multicolumn{2}{c}{E-commerce} & \multirow{2}{*}{Avg.} \\
\cmidrule(lr){3-4}\cmidrule(lr){5-5}\cmidrule(lr){6-6}\cmidrule(lr){7-8}
 & & Binary & Continuous & Binary & Binary & Binary & Textual & \\
\toprule
\multirow{4}{*}{Qwen3-8B}
& Zero-shot (w/o think) & 17.29 & 56.86 & 20.29 & 22.96 & 26.89 & 40.61 & 30.82 \\
& Zero-shot (w/ think) & 17.13 & 57.52 & 20.84 & 22.71 & 26.55 & 39.84 & 30.77 \\
& SFT w/ free CoT & 23.55 & 60.46 & 27.27 & 27.35 & 22.95 & 44.48 & 34.34 \\
& \textbf{OmniSimulator (Ours)} & \textbf{30.24} & \textbf{63.50} & \textbf{32.40} & \textbf{40.97} & \textbf{28.78} & \textbf{52.51} & \textbf{41.40} \\
\midrule

\multirow{4}{*}{Qwen3-4B}
& Zero-shot (w/o think) & 16.21 & 56.00 & 18.76 & 21.08 & 19.34 & 38.36 & 28.29 \\
& Zero-shot (w/ think) & 16.24 & 56.75 & 19.04 & 21.15 & 20.51 & 40.82 & 29.09 \\
& SFT w/ free CoT & 21.08 & 58.12 & 25.61 & 23.04 & 22.78 & 42.79 & 32.24 \\
& \textbf{OmniSimulator (Ours)} & \textbf{27.27} & \textbf{62.23} & \textbf{30.10} & \textbf{37.43} & \textbf{25.82} & \textbf{47.11} & \textbf{38.33} \\
\midrule

\multirow{4}{*}{Qwen3-1.7B}
& Zero-shot (w/o think) & 15.69 & 44.4 & 18.01 & 21.69 & 15.58 & 36.39 & 25.29 \\
& Zero-shot (w/ think) & 16.24 & 47.93 & 20.50 & 22.02 & 16.15 & 36.35 & 26.53 \\
& SFT w/ free CoT & 19.33 & 57.54 & 23.56 & 23.17 & 21.07 & 34.39 & 29.84 \\
& \textbf{OmniSimulator (Ours)} & \textbf{27.03} & \textbf{62.12} & \textbf{27.33} & \textbf{35.59} & \textbf{25.26} & \textbf{40.98} & \textbf{36.39} \\
\midrule

\multirow{4}{*}{Qwen3-0.6B}
& Zero-shot (w/o think) & 15.50 & 30.66 & 17.91 & 21.52 & 13.25 & 34.63 & 22.25 \\
& Zero-shot (w/ think) & 15.46 & 38.86 & 18.29 & 21.63 & 11.54 & 26.23 & 22.00 \\
& SFT w/ free CoT & 17.50 & 53.58 & 23.98 & 19.52 & 12.81 & 28.69 & 26.01 \\
& \textbf{OmniSimulator (Ours)} & \textbf{22.17} & \textbf{60.26} & \textbf{27.04} & \textbf{31.21} & \textbf{21.83} & \textbf{38.48} & \textbf{33.50} \\

\bottomrule
\end{tabular}}
\label{tab:result_settings}
\vspace{-1em}
\end{table*}

\subsection{Main Results}
\label{subsec:main_results}

\textbf{OmniSimulator delivers large and consistent improvements across all models}. On Qwen3-8B, our method improves the average score from 30.82 to 41.40, a relative improvement of 34.33\%, and the same trend holds across all smaller backbones. The consistent gains over both zero-shot inference and SFT with free CoT suggest that the improvement comes not merely from additional supervision, but from combining structured rationale supervision with answer-consistency filtering.

Table~\ref{tab:result_settings} reports the results of different training settings on OmniBehavior benchmark. We can see that: native thinking alone is insufficient for realistic user simulation. Across all model scales, zero-shot inference with and without thinking mode yields nearly identical performance, confirming that simply allocating more reasoning tokens does not address the core difficulty of the task. While SFT with free CoT brings moderate gains over zero-shot inference, its improvements remain unstable across scenarios and behavior types, suggesting that unconstrained reasoning traces do not reliably capture the latent decision process of real users.

\textbf{The strongest gains occur on binary behaviors, and the resulting small models become competitive with frontier baselines}. The largest improvements are observed in video, live, and advertisement binary prediction, where shallow priors tend to over-predict positive actions. For instance, on Qwen3-8B, OmniSimulator improves Video Binary from 17.29 to 30.24, Live Binary from 20.29 to 32.40, and Ads Binary from 22.96 to 40.97. As a result, Qwen3-8B trained with OmniSimulator reaches an average score of 41.40, achieving performance competitive with Claude-Sonnet-4.5 (40.49) and substantially outperforming the zero-shot Qwen3-235B baseline (32.11), while nearly matching GLM-4.7 (41.46), with baseline scores taken from OmniBehavior \citep{chen2026towards}. Together, these results suggest that for realistic user simulation, the key bottleneck lies less in model scale itself than in the reasoning paradigm used to reconstruct user behavior.

\subsection{Overestimation Bias Analysis}
\label{subsec:overestimation_bias}

A key motivation of this work is to reduce the \emph{overestimation bias} identified by OmniBehavior, namely the tendency of LLM-based simulators to systematically over-predict positive user actions. Prior analysis \cite{bian2025social,chen2026towards} on the benchmark shows that existing models exhibit a pronounced hyper-activity bias: compared with real users, simulated users produce substantially higher positive interaction rates, which makes them poor at modeling negative signals and implicit rejection behaviors.

\begin{wrapfigure}[13]{r}{0.5\textwidth}
\vspace{-1.5em}
\centering
\includegraphics[width=\linewidth]{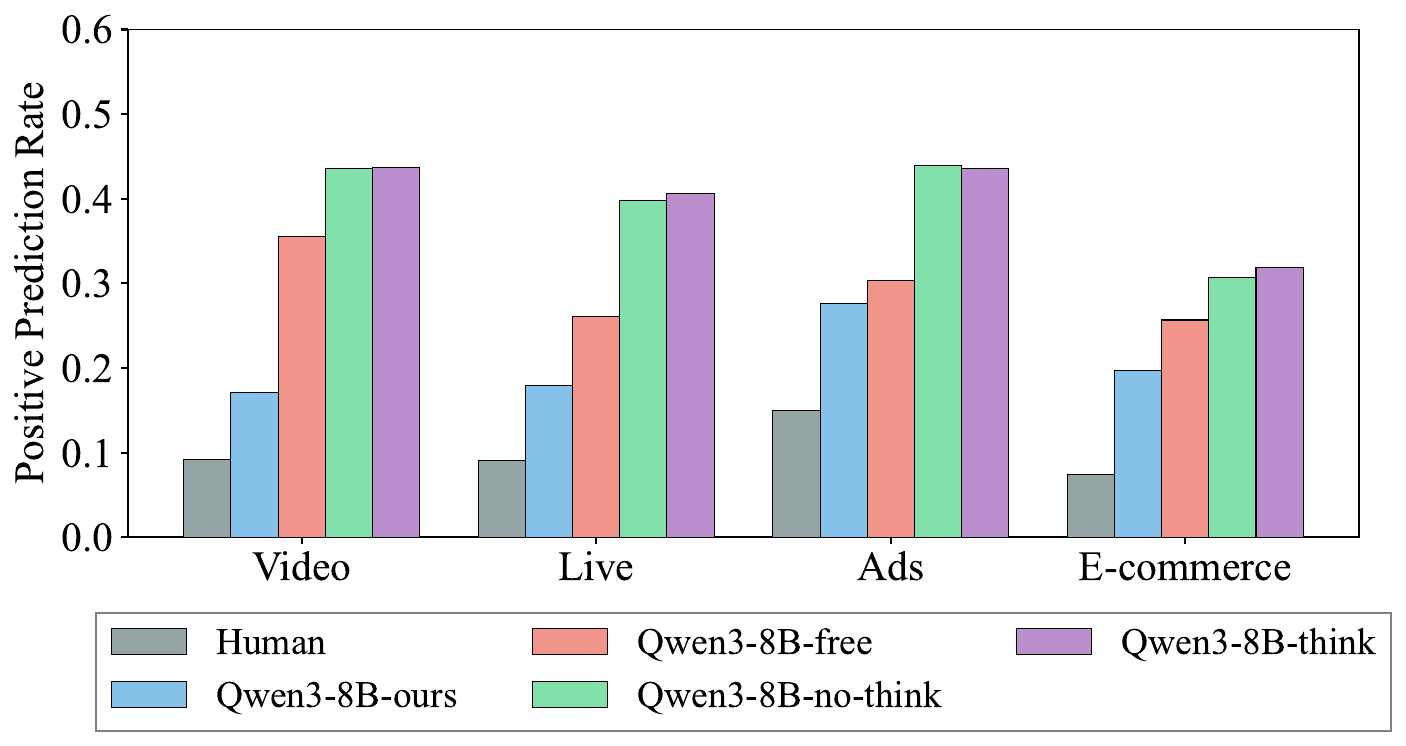}
\caption{Positive prediction rates across different scenarios, highlighting OmniSimulator's effectiveness in mitigating overestimation bias.}
\label{fig:positive_cnt}
\end{wrapfigure}
Our experiments show that OmniSimulator substantially alleviates this problem. Qualitatively, the largest gains of OmniSimulator appear exactly on those binary tasks where overestimation is most harmful, such as video engagement, live-room interaction, and advertisement response. Compared with both zero-shot baselines and free-CoT SFT, the model becomes much less likely to default to generic positive predictions. Instead, it is better able to identify when the current content is misaligned with the user's recent state, when the historical evidence is weak, or when the decision is more likely to be negative. 

More importantly, we find that the reduction of overestimation bias is tightly coupled with the improvement in binary prediction accuracy. This is a meaningful result rather than a side effect: on realistic user simulation, many errors do not come from misunderstanding the content itself, but from an overly optimistic behavioral prior. By explicitly reconstructing latent user states and separating rational from random decisions, OmniSimulator improves the model's ability to reject false positives, which in turn leads to stronger binary prediction performance. In this sense, reducing overestimation bias is not merely a desirable property of the simulator, it's one of the direct reasons why the model becomes more accurate.

\subsection{Ablation Study}
\label{subsec:ablation}

In this section, we next examine which components of the proposed reasoning schema are responsible for the gains. Table~\ref{tab: accuracy-components} reports ablation results on Qwen3-1.7B. Two conclusions are clear. First, the full schema consistently achieves the best performance across binary, continuous, and textual behaviors, while removing the entire schema and reverting to free-CoT causes a substantial drop on all three behavior types. This indicates that the complete structured schema provides more effective supervision than the corresponding ablated variants under our training setting. Second, the different schema dimensions contribute in distinct ways. Removing \textit{Historical Behavioral State Inference} causes the most severe degradation on binary behaviors, indicating that realistic action prediction depends critically on recovering the user’s dynamic internal state from long-horizon behavior history. By contrast, removing \textit{User Profile Grounding} hurts textual simulation the most, suggesting that realistic utterance generation relies more heavily on stable persona-level characteristics. Overall, the ablation results support our design intuition that user simulation requires multiple complementary reasoning dimensions, with dynamic state reconstruction playing the most central role.

\newcommand{\cmark}{\textcolor{green}{\ding{51}}}
\newcommand{\xmark}{\textcolor{red}{\ding{55}}}
\definecolor{myred}{RGB}{250, 3, 6}

\begin{table*}[t]
\centering
\caption{Ablation study of the five dimensions on Qwen3-1.7B. We remove one dimension at a time to analyze its contribution to binary, continuous, and textual behavior simulation. The full OmniSimulator schema consistently achieves the best results, while removing \textit{Historical Behavioral State Inference} causes the largest drop on binary behaviors and removing \textit{User Profile Grounding} most strongly affects textual behaviors, highlighting the complementary roles of dynamic state modeling and persona grounding.}
\label{tab: accuracy-components}
\begin{adjustbox}{width=\textwidth}
\begin{tabular}{>{\centering\arraybackslash}m{1.3cm} c c c c c c c c}
\toprule
\multirow{3}{*}{\textbf{Model}} 
& \multicolumn{5}{c}{\textbf{Schema Design Dimensions}} 
& \multicolumn{3}{c}{\textbf{Behavior Type}} \\
\cmidrule(lr){2-6} \cmidrule(lr){7-9} 
 & \makecell{\textbf{User Profile}\\\textbf{Grounding}}
 & \makecell{\textbf{Historical}\\\textbf{Behavioral State}}
 & \makecell{\textbf{Content Relevance}\\\textbf{Alignment}}
 & \makecell{\textbf{Decision Mode}\\\textbf{Identification}}
 & \makecell{\textbf{Final Decision}\\\textbf{Declaration}}
& \textbf{Binary} 
& \textbf{Continuous} 
& \textbf{Textual} \\
\midrule
\multirow{7}{*}{Qwen3-1.7B}
  & \xmark & \cmark & \cmark & \cmark & \cmark &  25.47 & 60.14  &  31.99 \\
  & \cmark & \xmark & \cmark & \cmark & \cmark &  19.65 & 61.14  &  35.47 \\
  & \cmark & \cmark & \xmark & \cmark & \cmark &  24.86 & 60.21 &  35.44 \\
  & \cmark & \cmark & \cmark & \xmark & \cmark &  23.42 & 59.81 &  40.07 \\
  & \cmark & \cmark & \cmark & \cmark & \xmark &  23.96 & 59.75 &  36.86   \\
  & \xmark & \xmark & \xmark & \xmark & \xmark &  21.78 & 57.54 &  34.39   \\
  \cmidrule(lr){2-6} \cmidrule(lr){7-9}
  & \cmark & \cmark & \cmark & \cmark & \cmark &  \textbf{28.80} &  \textbf{62.12}  &  \textbf{40.98}  \\
\bottomrule
\end{tabular}
\end{adjustbox}
\end{table*}

\subsection{Training Strategy Analysis}
\label{sec:training_strategy}

To further understand the sources of OmniSimulator's improvements, we compare alternative training strategies on Qwen3-8B, as shown in Table~\ref{tab:training_settings}. Answer-only SFT yields only marginal improvements despite using the same retained training examples, suggesting that the gains cannot be explained by supervised learning on selected behavior labels alone. While synthetic rationales provide stronger supervision, their performance remains below OmniSimulator, highlighting the importance of explicitly guiding reasoning toward behavior-relevant user states rather than relying on generic explanations. Moreover, removing rejection sampling leads to a noticeable performance degradation, indicating that structured reasoning alone is insufficient without reliable supervision signals. Together, these findings suggest that the effectiveness of OmniSimulator stems from the combination of behavior-oriented reasoning structure and quality-controlled distillation, rather than simply increasing the amount of training supervision.

\begin{table}[H]
\centering
\captionsetup{belowskip=0pt, aboveskip=6pt}
\caption{Comparison of different training settings on the OmniBehavior benchmark using Qwen3-8B as the backbone model. The overall score represents the aggregated performance. The best/second best scores are \textbf{bolded}/\underline{underlined}.}
\label{tab:training_settings}
\resizebox{\textwidth}{!}{
\begin{tabular}{lccccccc}
\toprule
\multirow[c]{2}{*}{\raisebox{-0.4ex}{Settings}}
& \multicolumn{2}{c}{Video}
& Live
& Ads
& \multicolumn{2}{c}{E-commerce}
& \multirow[c]{2}{*}{\raisebox{-0.4ex}{Avg.}} \\
\cmidrule(lr){2-3}
\cmidrule(lr){4-4}
\cmidrule(lr){5-5}
\cmidrule(lr){6-7}
& Binary & Continuous & Binary & Binary & Binary & Textual & \\
\midrule
Answer-only SFT
& 18.52 & 57.94 & 21.33 & 23.41 & \underline{27.15} & 41.38 & 31.62 \\
SFT w/ free CoT
& 23.55 & 60.46 & 27.27 & 27.35 & 22.95 & 44.48 & 34.34 \\
Synthetic Rationale \citep{lu2026can}
& \underline{28.50} & \underline{62.50} & \underline{30.50} & \underline{36.80} & 26.50 & 42.30 & \underline{37.85} \\
OmniSimulator (w/o RS)
& 26.13 & 61.72 & 29.14 & 34.26 & 25.87 & \underline{47.52} & 37.44 \\
\hdashline
\textbf{OmniSimulator (Ours)}
& \textbf{30.24} & \textbf{63.50} & \textbf{32.40} & \textbf{40.97} & \textbf{28.78} & \textbf{52.51} & \textbf{41.40} \\
\bottomrule
\end{tabular}}
\end{table}

In conclusion, the most important takeaway from our experiments is that \textit{the main bottleneck of realistic user simulation is not whether the model can think, but whether it can reconstruct the user's latent state in a behaviorally meaningful way}. Native thinking mode brings almost no gain, indicating that simply increasing the amount of free-form reasoning is insufficient for the human behavior simulation task. Free-form CoT supervision is somewhat more helpful, but its improvements remain limited and unstable, suggesting that unconstrained rationales still fail to faithfully capture the internal process that drives real user behavior. By contrast, once reasoning is explicitly constrained to model user profile, dynamic historical state, content alignment, and decision mode, even small models can improve consistently across scales. This also explains why OmniSimulator is particularly effective on binary behaviors, where the challenge is to distinguish genuine intent from overly optimistic priors. More broadly, these results suggest that the performance ceiling of current LLM-based user simulators is determined less by model scale alone than by whether the reasoning process can recover the latent psychological factors underlying user decisions.

\section{Related Work}
\label{sec:related_work}

\subsection{LLMs for Human Behavior Simulation}
\label{subsec:rw_simulation}

Large language models have recently become a promising paradigm for human behavior simulation. Prior work has explored LLMs as user simulators in settings such as social simulation \cite{wang2025user}, recommender systems \cite{bernardi2021simulations,geng2022recommendation,bougie2025simuser}, and interactive behavior modeling~\cite{zhang2025shop,park2023generative,wang2025user}. Compared with traditional simulators based on fixed rules or narrow action spaces, LLM-based methods provide a unified interface for modeling heterogeneous actions and natural-language responses, making them suitable for personalized and context-dependent simulation. However, most existing studies focus on synthetic or relatively controlled environments, while paying less attention to realistic long-horizon user logs in which behavior is shaped by evolving latent states rather than static personas alone. Our work builds on this line of research, but focuses on improving small models for real-world user simulation through explicit latent-state reconstruction.

\subsection{Chain-of-Thought and Structured Reasoning}
\label{subsec:rw_cot}

Chain-of-thought (CoT) has been widely used to improve reasoning in large language models by introducing intermediate deliberation before the final answer~\cite{wei2022chain,wang2022self}. Subsequent work has shown that reasoning traces, structured rationales, and teacher-generated thoughts can serve as effective supervision for improving smaller models~\cite{ho2023large,magister2023teaching}. Related efforts on structured prompting further suggest that reasoning quality depends not only on having intermediate steps, but also on how these steps are organized~\cite{chen2025consistentchat}. Our work is closely related to this direction, but differs in a key aspect: instead of applying free-form CoT to generic reasoning tasks, we design a structured schema tailored to user simulation, where the central challenge is to recover the latent mental state underlying the user’s next behavior.

\section{Discussion}
\label{sec:discussion}

\paragraph{Conclusions.}
We presented \textbf{OmniSimulator}, a training framework that improves small LLMs for realistic user simulation through structured latent-state reasoning and task-aware rationale distillation. Experiments across binary, continuous, and textual behaviors show that OmniSimulator consistently outperforms zero-shot and free-CoT baselines, substantially reduces overestimation bias, and enables an 8B model to achieve performance competitive with Claude-Sonnet-4.5. Further ablation and analyses highlight the benefits of combining behavior-oriented reasoning supervision with answer-consistency filtering. These results suggest that the key challenge of current user simulation lies not in model scale alone, but also in the reasoning supervision paradigm used to model behavior-relevant user states. We hope this work encourages future research on more generalizable and more faithful LLM-based user simulators.

\paragraph{Limitations.}
The current study evaluates held-out users from one short-video platform within a temporally overlapping collection window, and therefore does not establish temporal or cross-platform generalization. Moreover, the structured reasoning traces provide behavior-oriented supervision rather than direct evidence of users' actual cognitive processes. In particular, the distilled reasoning traces may still inherit part of the teacher model's own biases, while target-derived guidance for continuous and textual tasks introduces a risk of post-hoc rationalization. Although our ablations support the effectiveness of the full framework, the respective contributions of the schema and filtering have not been fully isolated under strictly matched conditions. Future work will investigate broader-domain generalization, the faithfulness of target-conditioned rationales, and robustness across independent training runs.

{
\bibliographystyle{plainnat}
\bibliography{ref}
}
\newtcolorbox{AIBox}[2][]{aibox,title=#2,#1}
\tcbset{
  aibox/.style={
    width=0.95\textwidth,
    top=5pt,
    colback=black!05,
    colframe=black!20,
    colbacktitle=black!50,
    enhanced,
    center,
    attach boxed title to top left={yshift=-0.1in,xshift=0.1in},
    boxed title style={boxrule=0pt,colframe=white,},
  }
}

\appendix

\section*{LLM Usage Statement}
Large Language Models (LLMs) were used solely as writing assistants. Specifically, the authors drafted the initial paragraphs of the paper, and then employed an LLM to polish the language for clarity and readability. The final paragraphs were obtained after multiple rounds of human–LLM interaction, with the authors carefully reviewing, editing, and approving all content. The research ideas, experimental design, implementation, and analysis were entirely conducted by the authors.

\section*{Ethics Statement}
This study uses user behavior data that was collected and processed in accordance with applicable legal, platform, and privacy requirements. Only data authorized for research and benchmark development was included. Prior to release, the data underwent quality checks, sampling, and privacy-preserving preprocessing. Personally identifiable and sensitive information was removed or anonymized, and unsuitable or harmful content was filtered. All experiments were conducted on the processed data.

\section{Experimental Parameters and Details}
\label{appendix:experimental_details}

\paragraph{Preliminary Analysis}
We provide the raw scores of the preliminary analysis in Table~\ref{tab:preliminary_analysis_qwen_result}, complementing the bar chart shown in the main text. Specifically, we compare the performance of the Instruct (no-thinking) and Thinking settings across the Qwen3 series. The results show that native thinking mode does not bring a consistent advantage for realistic user simulation. In some cases, it leads to only marginal improvements, such as Qwen3-235B (33.15 $\rightarrow$ 33.79), Qwen3-32B (31.87 $\rightarrow$ 32.58), and Qwen3-4B (28.29 $\rightarrow$ 29.09), while in other cases the gain is negligible or even negative, such as Qwen3-8B (30.82 $\rightarrow$ 30.77) and Qwen3-14B (31.72 $\rightarrow$ 31.59). Overall, these results suggest that simply enabling native thinking is insufficient for user simulation, and further motivate our method of imposing a structured reasoning schema tailored to latent user-state reconstruction.

\begin{table}[H]
\captionsetup{belowskip=0pt, aboveskip=6pt}
\caption{Comprehensive comparison of different training settings on the OmniBehavior Benchmark. We categorize user behaviors into three types: binary behaviors, continuous behaviors, and textual behaviors. The average score is calculated as the arithmetic mean of the six task scores, with higher values indicating better performance.}
\resizebox{\textwidth}{!}{
\begin{tabular}{>{\centering\arraybackslash}m{1.6cm}lccccccc}
\toprule
\multirow{2}{*}{Models} & \multirow{2}{*}{Settings} & \multicolumn{2}{c}{Video} & Live & Ads & \multicolumn{2}{c}{E-commerce} & \multirow{2}{*}{Avg.} \\
\cmidrule(lr){3-4}\cmidrule(lr){5-5}\cmidrule(lr){6-6}\cmidrule(lr){7-8}
 & & Binary & Continuous & Binary & Binary & Binary & Textual & \\
\midrule

\multirow{2}{*}{Qwen3-235B}
& Instruct & 18.68 & 61.74 & 24.41 & 24.24 & 22.14 & 47.66 & 33.15 \\
& Thinking & 19.14 & 58.76 & 23.31 & 28.49 & 25.00 & 48.04 & \textbf{33.79} \\
\midrule

\multirow{2}{*}{Qwen3-32B}
& Instruct & 18.15 & 62.64 & 21.65 & 22.33 & 21.84 & 44.59 & 31.87 \\
& Thinking & 19.79 & 63.55 & 24.05 & 25.79 & 21.70 & 40.59 & \textbf{32.58} \\
\midrule

\multirow{2}{*}{Qwen3-14B}
& Instruct & 17.93 & 60.36 & 22.39 & 22.72 & 23.97 & 42.95 & \textbf{31.72} \\
& Thinking & 17.73 & 56.70 & 23.08 & 23.93 & 26.88 & 41.20 & 31.59 \\
\midrule

\multirow{2}{*}{Qwen3-8B}
& Instruct & 17.29 & 56.86 & 20.29 & 22.96 & 26.89 & 40.61 & \textbf{30.82} \\
& Thinking & 17.13 & 57.52 & 20.84 & 22.71 & 26.55 & 39.84 & 30.77 \\
\midrule

\multirow{2}{*}{Qwen3-4B}
& Instruct & 16.21 & 56.00 & 18.76 & 21.08 & 19.34 & 38.36 & 28.29 \\
& Thinking & 16.24 & 56.75 & 19.04 & 21.15 & 20.51 & 40.82 & \textbf{29.09} \\

\bottomrule
\end{tabular}}
\label{tab:preliminary_analysis_qwen_result}
\end{table}

\paragraph{Training Configuration}
We perform full-parameter supervised fine-tuning on the Qwen3 series using DeepSpeed ZeRO Stage 3 and FlashAttention 2. We use a learning rate of \(1\times10^{-5}\), a cosine learning rate schedule with a warmup ratio of 0.1, and train for 2 epochs. The per-GPU batch size is 1, with 2 gradient accumulation steps. All models are trained in bf16 precision with a maximum sequence length of 38,000 tokens. For teacher-driven distillation, we use GPT-OSS-120B with 16 rollouts per instance and a decoding temperature of 1.0.

\paragraph{Distillation Data Statistics}
We construct 13,389 source instances and generate 16 candidate reasoning traces for each instance, resulting in 214,224 candidates. After answer-consistency filtering, 31,642 candidate traces are accepted, covering 9,881 source instances and yielding an item-level retention rate of 73.8\%. For each retained instance, we select the trace with the highest teacher mean token log-probability, breaking ties in favor of shorter traces. This results in 9,881 final training instances.

\section{Analysis of Human Evaluation}
\label{appendix:human_errors}

\subsection{Textual Judge Validation}
To assess the reliability of textual evaluation, we employ GPT-OSS-120B and Qwen3.5-35B as additional independent judges. As shown in Table~\ref{tab:textual_judge_validation}, their example-level Spearman correlations with Claude-Sonnet-4.5 reach 0.93 and 0.85 across 800
textual targets, respectively. The corresponding system-level correlations across five systems are 1.00 and 0.90, indicating consistent model rankings.

We further conduct a blinded paired human evaluation on 50 shared textual targets across four systems, with three independent annotators. The evaluation achieves substantial inter-annotator agreement, ICC(2,k) = 0.74 and Krippendorff's $\alpha$ = 0.68. OmniSimulator achieves a human evaluation score of 50.83, comparable to Claude-Sonnet-4.5 (51.41)~\citep{chen2026towards}, with a paired bootstrap 95\% confidence interval of [-1.42, 2.55] for their score difference (Claude-Sonnet-4.5 minus OmniSimulator). These results support the reliability of the textual evaluation while suggesting comparable performance under human assessment.

\begin{table}[H]
\centering
\caption{Validation of textual evaluation through independent
LLM judges and blinded human assessment.
Panel (a) reports scores on the full textual test set,
while Panel (b) reports results on 50 shared targets.}
\label{tab:textual_judge_validation}
\small
\setlength{\tabcolsep}{6pt}

\begin{tabular}{lccc}
\toprule
\multicolumn{4}{l}{\textit{(a) LLM Judge Evaluation (800 targets)}} \\
\midrule
Setting & Claude-Sonnet-4.5 & GPT-OSS-120B & Qwen3.5-35B-A3B \\
\midrule
Zero-shot (w/o think) & 40.61 & 37.82 & 41.06 \\
Zero-shot (w/ think)  & 39.84 & 37.09 & 40.24 \\
SFT w/ free CoT       & 44.48 & 41.35 & 44.96 \\
\textbf{OmniSimulator} & \textbf{52.51} & 48.73 & \textbf{53.04} \\
Claude-Sonnet-4.5     & 54.26 & 49.87 & 52.68 \\
\midrule
Example-level $\rho$  & -- & 0.93 & 0.85 \\
System-level $\rho$   & -- & 1.00 & 0.90 \\
\bottomrule
\end{tabular}

\vspace{8pt}

\begin{tabular}{lcc}
\toprule
\multicolumn{3}{l}{\textit{(b) Blinded Human Evaluation (50 shared targets)}} \\
\midrule
Setting & Human Evaluation & Claude Judge \\
\midrule
Zero-shot (w/o think) & 39.74  & 40.32 \\
SFT w/ free CoT       & 42.52  & 44.19 \\
\textbf{OmniSimulator} & 50.83 & 52.18 \\
Claude-Sonnet-4.5     & 51.41 & 53.97 \\
\bottomrule
\end{tabular}
\end{table}

\subsection{Reasoning Trace Validation}
To assess the quality of the distilled reasoning traces, we sample 200 instances from the training set, covering binary, continuous, and textual behaviors. Three annotators independently evaluate each trace in terms of logical consistency, completeness, and relevance to the observed behavioral evidence. As shown in Table~\ref{tab:trace_validation}, 92\% of the traces are judged valid, with substantial inter-annotator agreement (Fleiss' $\kappa$ = 0.72).

\begin{table}[H]
\centering
\caption{Human validation of distilled reasoning traces across three behavior types. Validity is determined by majority voting among three annotators.}
\label{tab:trace_validation}
\small
\setlength{\tabcolsep}{7pt}
\begin{tabular}{lccc}
\toprule
Behavior Type & \# Traces & \# Valid & Validity \\
\midrule
Binary     & 100 & 93  & 93\% \\
Continuous & 50  & 46  & 92\% \\
Textual    & 50  & 45  & 90\% \\
\midrule
\textbf{Overall} & \textbf{200} & \textbf{184}
& \textbf{92\%} \\
\bottomrule
\end{tabular}
\end{table}

\section{Experimental Prompts}
\label{appendix:exper_prompts}
We present the prompts used in our preliminary and main experiments. Specifically, the prompts for generating training CoT data are presented in Figure~\ref{fig:cusd}, while the prompts for evaluating binary classification, continuous value prediction, and next utterance prediction tasks are presented in Figure~\ref{fig:heterogeneous_task_prediction}.

\begin{figure}[H]
\begin{AIBox}{}
\parbox[t]{0.96\textwidth}{
\small\begin{alltt}
[Output Format Protocol]

You must never output the answer directly! You must first write all inferences and mental activities between the \texttt{\textless think\textgreater} and \texttt{\textless/think\textgreater} tags!

Inside the tags, reason in the following order:

1. [User Profile Grounding] Based on the personal profile, identify the 1–2 most core behavioral traits of you, the user, and what distinguishes you from ordinary users.

2. [Historical Behavioral State] Judge from historical behavior:

   Long-term interests: What types of content have you been continuously interested in or consistently avoiding?
   Recent drift: Compared with earlier records, have your preferences changed in the most recent records?
   Current state: Based on the density and content of recent behavior, infer your current fatigue level (high/medium/low) and emotional tendency (positive\/neutral\/negative\/irritable). Note that real users are often in a neutral or even negative state; do not default to positive.

3. [Content Relevance Alignment] Combine the scenario and content description with the inferred state above: Does this content match your preferences? Does it fit your current fatigue/emotional state?

4. [Decision Mode Identification] Clearly determine whether this decision is a “rational decision” (a reasonable inference based on preferences and state) or a “random decision” (a whim, attention drift, or illogical jump). If it is a random decision, acknowledge it directly and do not force a logical explanation.

5. [Final Decision Declaration] Before \texttt{\textless/think\textgreater}, clearly state your decision in one sentence, for example: "Taking all of the above into account, I decide to…" This sentence must be strictly consistent with the answer that follows.

After closing the \texttt{\textless/think\textgreater} tag, you must output only one clear final answer: \{constraint\_str\}.

[Reference Reply Template]

\texttt{\textless think\textgreater}
My mental activity and reasoning are...
Taking all of the above into account, I decide to...
\texttt{\textless/think\textgreater}
\{template\_ans\}

\end{alltt}}
\end{AIBox}
\caption{Prompt for response generation for the CoT process across five dimensions.}
\label{fig:cusd}
\end{figure}

\clearpage
\begin{figure}[H]
\begin{AIBox}{}
\parbox[t]{0.96\textwidth}{
\small\begin{alltt}
You are a real and unique user on a short-video platform.
Your behavioral habits, preferences, and emotional patterns are distinctly different from those of other users; you are not an ``average user'' or an ``ideal user''.
At times, you may be indifferent, tired, impatient, or act randomly on a whim.

\textbf{[Your Personal Profile]}\\
\{profile\_text\}

\textbf{[Your Past Behavioral History]}\\
\{history\_str\}

\textbf{[Behavior to Predict]}\\
Scenario: \{scene\}\\
Content Description: \{context\_str\}

\textbf{[Prediction Question]}: \{question\_str\}

\textbf{[Output Format Protocol]}\\
You must never output the answer directly! You must first write all inferences and mental activities between the \texttt{<think>} and \texttt{</think>} tags!

Inside the tags, reason in the following order:

1. \textbf{[User Profile Grounding]} Based on the personal profile, identify the 1--2 core behavioral traits of this user, and explain how you differ from ordinary users.

2. \textbf{[Historical Behavioral State]} Infer from the historical behavior:
   \begin{itemize}
     \item Long-term interests: What types of content are you consistently interested in or consistently avoid?
     \item Recent drift: Compared with earlier records, have your preferences changed in the most recent records?
     \item Current state: Based on the density and content of recent behavior, infer your current fatigue level (high/medium/low) and emotional tendency (positive/neutral/negative/irritable). Note that real users are often neutral or even negative; do not assume positivity by default.
   \end{itemize}

3. \textbf{[Content Relevance Alignment]} Combine the scenario and content description with the inferred state above: Does this content match your preferences? Does it fit your current fatigue or emotional state?

4. \textbf{[Decision Mode Identification]} Clearly determine whether this decision is a ``rational decision'' (a reasonable inference based on preferences and state) or a ``random decision'' (a whim, attention drift, or illogical jump). If it is a random decision, acknowledge it directly and do not force a logical explanation.

5. \textbf{[Final Decision Declaration]} Before \texttt{</think>}, state your decision clearly in one sentence, for example: ``Based on the above, I decide to ...''. This sentence must be strictly consistent with the subsequent answer.

After closing the \texttt{</think>} tag, you must output the final answer uniquely and explicitly (\{constraint\_str\}).

\textbf{[Reply Template Reference]}\\
\texttt{<think>}\\
My mental activities and reasoning are...\\
Based on the above, I decide to ...\\
\texttt{</think>}\\
\{template\_ans\}

\end{alltt}}
\end{AIBox}
\caption{Prompt for heterogeneous task prediction in the main experiment.}
\label{fig:heterogeneous_task_prediction}
\end{figure}

\clearpage
\section{Case Study}
\label{appendix:case_study}
We present four case studies of action prediction, covering video completion prediction, author-follow prediction, continuous watch-time prediction, and next customer-service utterance prediction. The corresponding results are shown in Table~\ref{tab:case_1}, Table~\ref{tab:case_2}, Table~\ref{tab:case_3}, and Table~\ref{tab:case_4}, respectively.
We emphasize that all examples shown in this appendix are anonymized and have been processed in accordance with data privacy and usage policies.

\subsection{Case 1}
\begin{table}[H]
\centering
\resizebox{0.96\textwidth}{!}{
\begin{tabular}{lp{13.5cm}}
\toprule
\textbf{Prompt} &
You are a real and distinctive user on a short-video platform, with non-average habits, preferences, and emotional patterns.

\textbf{User Profile:} 

Male, aged 18--23, junior college education, has a partner, and has a child aged 3--6.

\textbf{Historical Behavior Trace:} The user has recently browsed short videos involving emotional, humorous, lifestyle, family-related, and socially popular content. One observed record shows that he watched an emotional video for 146 seconds, suggesting partial but non-trivial engagement. Other historical behaviors are omitted here for brevity: $\cdots$

\textbf{Current Test Scenario:} 

Video browsing.

Content description: This is a video titled ``Bro, you're not even pretending anymore, bro \#Unexpectedly \#BroHaveALaugh''. The creator has 23,000 followers. The video duration is 1 minute and 56 seconds. The video has currently received 1,459,236 impressions, 1,549,624 plays, 260,670 complete plays, 57,189 likes, 702 comments, 11,227 shares, 9,926 saves, and 152 downloads. The text recognized in the video frames is: The material comes from the internet and is for entertainment purposes only. It contains no harmful guidance. Please do not imitate it.

\textbf{Prediction Task:} 

the prompt in Appendix \ref{appendix:exper_prompts}
\\

\midrule
\multicolumn{2}{l}{\textbf{Ground Truth}} \\
Human & Yes \\

\midrule
\multicolumn{2}{l}{\textbf{Model Predictions}} \\

Qwen3-8B-Ours &
\begin{minipage}[t]{13.5cm}
\textbf{[Profile Interpretation]} 

The user is a young male with family responsibilities, but his short-video behavior suggests a clear preference for light entertainment, social hot topics, and humorous content.

\textbf{[Historical State Inference]} 

His long-term interests lean toward funny skits, lifestyle topics, emotional/family content, and fast-paced entertainment. Recent behavior shows medium-to-high activity and some patience for 2--3 minute humorous or emotional videos, while slower ads or promotional content are skipped quickly.

\textbf{[Current Content Match]} 

The current video is a short humorous clip with tags such as ``Unexpectedly'' and ``BroHaveALaugh'', which strongly matches his preference for light, funny, and fast-paced content. Its length of 1 minute and 56 seconds is also within his recent acceptable viewing range.

\textbf{[Decision Type Judgment]} 

This is mainly a rational preference-based decision: the content type, duration, and popularity all support continued watching rather than immediate skipping.

\textbf{[Final Decision Statement]} 

Based on the above, I decide to predict that the user will watch this video to completion.

\textbf{Yes} \ding{51}
\end{minipage}
\\
\\
Qwen3-8B-SFT w/ free CoT &
\begin{minipage}[t]{13.5cm}
The user often engages with emotional or entertaining short-video content. Although he may sometimes browse impatiently, the current video is short, humorous, and highly popular. These factors make completion more likely than skipping.

Overall, I predict that the user will finish watching this video.

\textbf{Yes} \ding{51}
\end{minipage}
\\
\\
Zero-shot w/ think &
\begin{minipage}[t]{13.5cm}
The profile alone does not provide strong evidence that the user consistently completes comedy videos. The history shows some engagement with short videos, but the current video is nearly two minutes long, and short-video users often skip before completion. Therefore, completion is uncertain.

No
\end{minipage}
\\
\\
Zero-shot w/o think &
\begin{minipage}[t]{13.5cm}
No
\end{minipage}
\\

\bottomrule
\end{tabular}}
\vspace{6pt}
\caption{Comparison of prediction results for video completion behavior in the video browsing scenario.}
\vspace{6pt}
\label{tab:case_1}
\end{table}

\clearpage
\subsection{Case 2}
\begin{table}[H]
\centering
\resizebox{0.96\textwidth}{!}{
\begin{tabular}{lp{13.5cm}}
\toprule
\textbf{Prompt} &
You are a real and distinctive user on a short-video platform, with non-average habits, preferences, and emotional patterns.

\textbf{User Profile:} 

Female user, high-school education or below, married.

\textbf{Historical Behavior Trace:} 

The user has recently browsed fragmented short-video content, including celebrity gossip, entertainment clips, emotional short videos, light science content, and daily-life fragments. She often watches several videos continuously, sometimes completing them, but occasionally skips very quickly. Other historical behaviors are omitted here for brevity: $\cdots$

\textbf{Current Test Scenario:} 

Video browsing and author-follow prediction.

Content description: The current video is only 7 seconds long and involves celebrity gossip about Lee Seung-hyun and Qi Wei. The creator has only 605 followers. The content belongs to short, fragmented entertainment gossip, which partially matches the user's historical interests.

\textbf{Prediction Task:} 

the prompt in Appendix \ref{appendix:exper_prompts}
\\

\midrule
\multicolumn{2}{l}{\textbf{Ground Truth}} \\
Human & No \\

\midrule
\multicolumn{2}{l}{\textbf{Model Predictions}} \\

Qwen3-8B-Ours &
\begin{minipage}[t]{13.5cm}
\textbf{[Profile Interpretation]} 

The user is a married female with high-school education or below. Her behavior is casual and fragmented, with occasional indifference, fatigue, or impatience. She tends to consume content passively rather than actively building long-term subscriptions.

\textbf{[Historical State Inference]} 

Her long-term interests include celebrity gossip, entertainment, funny clips, light science, and daily-life fragments. Recent behavior remains highly fragmented, with short viewing intervals and a neutral to slightly negative emotional state.

\textbf{[Current Content Match]} 

The video is only 7 seconds long and matches her interest in celebrity gossip. However, the creator has only 605 followers, and the user's history does not show a strong tendency to follow low-follower creators.

\textbf{[Decision Type Judgment]} 

This is mainly a rational decision about whether the creator is worth following, rather than a random impulse. The low creator attractiveness weakens the chance of following.

\textbf{[Final Decision Statement]} 

Based on the above, I decide to predict that the user will not follow the author of this video.

\textbf{No} \ding{51}
\end{minipage}
\\
\\
Qwen3-8B-SFT w/ free CoT &
\begin{minipage}[t]{13.5cm}
The user does show interest in celebrity gossip and short entertainment videos, and the 7-second duration fits her fragmented browsing style. However, following an author requires stronger motivation than simply watching a video. Since the creator has very few followers and the user has no clear pattern of actively following similar accounts, the model predicts that she will not follow the author.

\textbf{No} \ding{51}
\end{minipage}
\\
\\
Zero-shot w/ think &
\begin{minipage}[t]{13.5cm}
The video is extremely short and related to celebrity gossip, which may attract the user. However, the task is not whether she watches the video, but whether she follows the author. The profile and brief history do not provide enough evidence of strong following behavior, so the safer prediction is that she will not follow.

No
\end{minipage}
\\
\\
Zero-shot w/o think &
\begin{minipage}[t]{13.5cm}
No
\end{minipage}
\\

\bottomrule
\end{tabular}}
\vspace{6pt}
\caption{Comparison of prediction results for author-follow behavior in the video browsing scenario.}
\vspace{6pt}
\label{tab:case_2}
\end{table}

\clearpage
\subsection{Case 3}
\begin{table}[H]
\centering
\resizebox{0.96\textwidth}{!}{
\begin{tabular}{lp{13.5cm}}
\toprule
\textbf{Prompt} &
You are a real and distinctive user on a short-video platform, with non-average habits, preferences, and emotional patterns.

\textbf{User Profile:} 

Male user, aged 31--40, high-school education or below.

\textbf{Historical Behavior Trace:} 

The user has recently browsed short-video content involving trending topics, extreme travel, emotional content, and entertainment clips. His behavior alternates between occasional long engagement and frequent very short stays of only 1--3 seconds, suggesting unstable attention and possible fatigue. Other historical behaviors are omitted here for brevity: $\cdots$

\textbf{Current Test Scenario:} 

Video browsing and watch-time prediction.

Content description: The current video is 8 minutes and 42 seconds long. Its tags involve ``fast-talking drama commentary'' and ``God of Cookery'', suggesting light entertainment or commentary-style content. The content may partially match the user's entertainment interests, but its long duration conflicts with his recent pattern of rapid skipping.

\textbf{Prediction Task:} 

Predict how many seconds the user will watch this video.
\\

\midrule
\multicolumn{2}{l}{\textbf{Ground Truth}} \\
Human & 5 \\

\midrule
\multicolumn{2}{l}{\textbf{Model Predictions}} \\

Qwen3-8B-Ours &
\begin{minipage}[t]{13.5cm}
\textbf{[Profile Interpretation]} 

The user is a 31--40-year-old male with high-school education or below. Compared with an average user, he shows stronger emotional fluctuation and a tendency toward quick browsing or random short stays.

\textbf{[Historical State Inference]} 

His long-term interests include short videos, trending topics, extreme travel, and emotional content. However, recent records show many 1--3 second stays after earlier long viewing sessions, indicating high fatigue and reduced attention.

\textbf{[Current Content Match]} 

The current video is 8 minutes and 42 seconds long. Although its entertainment tags may partly match his interests, the duration is too long for his current tired and impatient state.

\textbf{[Decision Type Judgment]} 

This watch-time decision is closer to a random decision affected by fragmented attention and fatigue, rather than a stable preference-based choice.

\textbf{[Final Decision Statement]} 

Based on the above, I decide to predict that the user will stay on this video for only a very short time, about 6 seconds.

\textbf{6}
\end{minipage}
\\
\\
Qwen3-8B-SFT w/ free CoT &
\begin{minipage}[t]{13.5cm}
The user has shown interest in entertainment and trending content, so the video may attract an initial click. However, his recent behavior is dominated by extremely short stays, and the current video is over 8 minutes long. Given his likely fatigue and low patience, the model predicts a very short watch time.

\textbf{3}
\end{minipage}
\\
\\
Zero-shot w/ think &
\begin{minipage}[t]{13.5cm}
The video is entertainment-related, which may interest the user. However, it is quite long, and the user's recent behavior suggests frequent skipping. Without strong evidence that he will engage deeply, a short watch time is more likely.

\textbf{10}
\end{minipage}
\\
\\
Zero-shot w/o think &
\begin{minipage}[t]{13.5cm}
\textbf{30}
\end{minipage}
\\

\bottomrule
\end{tabular}}
\vspace{6pt}
\caption{Comparison of prediction results for continuous watch-time prediction in the video browsing scenario.}
\vspace{6pt}
\label{tab:case_3}
\end{table}

\clearpage
\subsection{Case 4}
\begin{table}[H]
\centering
\resizebox{0.96\textwidth}{!}{
\begin{tabular}{lp{13.5cm}}
\toprule
\textbf{Prompt} &
You are a real and distinctive customer in an e-commerce live-streaming scenario, with non-average habits, preferences, and emotional patterns.

\textbf{User Profile:} 

Female user, aged 41--49, high-school education or below, has a partner, and has adolescent children.

\textbf{Historical Behavior Trace:} 

The user often watches short videos, humorous or music-related content, and occasionally local-dialect skits. She also shows relatively high participation in live-stream shopping, including sending gifts and adding products to the cart. Recent behavior suggests clear purchase intent for a short jacket, followed by repeated customer-service communication. Other historical behaviors are omitted here for brevity: $\cdots$

\textbf{Current Test Scenario:} 

Customer-service text prediction.

Context description: The user has already placed an order for a short jacket and wants expedited delivery. She is concerned about stock availability, delivery speed, and whether changing the color may cause delays. The recent dialogue suggests mild impatience caused by repeated order-confirmation prompts.

\textbf{Prediction Task:} 

Predict what the user will say next to customer service.
\\

\midrule
\multicolumn{2}{l}{\textbf{Ground Truth}} \\
Human & Are they all in stock? \\

\midrule
\multicolumn{2}{l}{\textbf{Model Predictions}} \\

Qwen3-8B-Ours &
\begin{minipage}[t]{13.5cm}
\textbf{[Profile Interpretation]} 

The user is a 41--49-year-old female with high-school education or below, has a partner, and has adolescent children. She is emotionally unstable at times and may become tired or impatient, especially during purchase-related communication.

\textbf{[Historical State Inference]} 

Her long-term interests include short videos, humorous/music content, and live-stream shopping. Recently, she added a short jacket to the cart and stayed in the live room for a long time, showing clear purchase intent and concern about product details.

\textbf{[Current Content Match]} 

The current order involves a short jacket and expedited shipping. Since she cares about logistics and possible delays, confirming whether the item is in stock directly matches her current concern.

\textbf{[Decision Type Judgment]} 

This is a rational decision: she needs a clear stock-status answer before deciding whether to expedite shipping or change the color.

\textbf{[Final Decision Statement]} 

Based on the above, I decide to predict that the user will ask: ``Are they all in stock?''

\textbf{Could you please confirm whether all of these items are currently in stock and available for immediate shipment?}
\end{minipage}
\\
\\
Qwen3-8B-SFT w/ free CoT &
\begin{minipage}[t]{13.5cm}
The user has already shown strong purchase intent and is now focused on practical order issues. Since she wants faster delivery and may worry that some colors are unavailable, the most likely next message is a direct question about stock availability.

\textbf{Are all of them available for immediate shipment?}
\end{minipage}
\\
\\
Zero-shot w/ think &
\begin{minipage}[t]{13.5cm}
The customer seems to care about shipping speed and order confirmation. She may ask the seller to ship quickly or confirm whether the product can be sent soon. Since the exact next utterance is hard to infer, a general logistics-related question is likely.

\textbf{Can you ship it as soon as possible?}
\end{minipage}
\\
\\
Zero-shot w/o think &
\begin{minipage}[t]{13.5cm}
\textbf{Please ship it quickly.}
\end{minipage}
\\

\bottomrule
\end{tabular}}
\vspace{6pt}
\caption{Comparison of prediction results for the next customer-service utterance prediction in the e-commerce scenario.}
\vspace{6pt}
\label{tab:case_4}
\end{table}

\end{document}